\documentclass[10pt,twocolumn,letterpaper]{article}

\usepackage{cvpr} 
\let\originalsection\section
\renewcommand{\section}{\vspace{-5pt}\originalsection}

\let\originalsubsection\subsection
\renewcommand{\subsection}{\vspace{-3pt}\originalsubsection}

\let\originalsubsubsection\subsubsection
\renewcommand{\subsubsection}{\vspace{-2pt}\originalsubsubsection}

\makeatother
\usepackage{times}
\usepackage{epsfig}
\usepackage{graphicx}
\usepackage{amsmath}
\usepackage{amssymb}
\usepackage{booktabs}
\usepackage{booktabs, multirow, makecell, array, siunitx}

\usepackage[most]{tcolorbox} 
\usepackage{xcolor}
\usepackage{colortbl}
\usepackage{graphicx}
\usepackage{caption}
\usepackage{algorithm}
\usepackage{algpseudocode}
\usepackage{tikz}
\usetikzlibrary{arrows.meta,positioning,fit,shapes.geometric,backgrounds}
\usepackage{booktabs}
\usepackage{multirow}
\usepackage{enumitem}
\usepackage{placeins}
\usepackage{amssymb}  
\usepackage{pifont}   
\usepackage{pifont} 
\usepackage{bm}
\usepackage[pagebackref,breaklinks,colorlinks]{hyperref}
\usepackage[table]{xcolor}
\usepackage{booktabs}
\usepackage{siunitx}
\usepackage{caption}
\definecolor{headerblue}{RGB}{220,230,242}
\definecolor{subheaderblue}{RGB}{235,242,250}
\definecolor{totalgray}{RGB}{245,245,245}
\usepackage[capitalize]{cleveref}
\crefname{section}{Sec.}{Secs.}
\Crefname{section}{Section}{Sections}
\Crefname{table}{Table}{Tables}
\crefname{table}{Tab.}{Tabs.}

\begin{document}
\title{
{\textcolor{red}{\small\bfseries Paper accepted at IEEE/IAPR International Joint Conference on Biometrics (IJCB), September 2026.}}\\[0.8em]
Detecting Clear Contact Lenses for Iris Recognition:\\ A Two-Stage Mask-Guided Attention Approach
}

\author{Parisa Farmanifard and Arun Ross\\
Michigan State University, East Lansing, MI 48824\\
{\tt\small \{farmanif,rossarun\}@msu.edu}
}

\maketitle
\thispagestyle{empty}

\begin{abstract}
This work focuses on the impact and detection of clear contact lenses in the context of iris recognition. While the detection of cosmetic or patterned contact lenses has been extensively studied under the presentation attack detection (PAD) paradigm, clear prescription contact lenses, that are typically transparent, have received comparatively less attention despite their widespread use. Unlike patterned lenses, clear lenses introduce no salient texture artifact, making them difficult to detect and are often assumed to have no impact on iris recognition. We first examine this assumption using the commercial VeriEye matcher on four benchmark datasets and show that clear lenses marginally degrade genuine match scores and increase verification error. We then propose a two-stage contact-lens detection framework. Stage~1 uses an existing PAD model to identify patterned lenses, while Stage~2 focuses on the more challenging clear-lens versus no-lens distinction using a ConvNeXt-Base model equipped with Mask-Guided Spatial Attention (MGSA). The proposed MGSA module incorporates a Hough-derived anatomical ROI mask together with learned spatial attention and Squeeze-and-Excitation channel recalibration, allowing the network to focus on subtle limbal cues associated with clear lens wear. Across four datasets, the full pipeline consisting of both patterned and clear contact lens detection achieves between 90.0\%--98.8\% accuracy. Finally, we introduce a z-score calibration method that adjusts VeriEye match scores when a clear lens is detected in the input images. This calibration reduces EER by 4.1\%--28.3\% across datasets, demonstrating that reliable clear contact lens detection can directly improve iris verification performance.
\end{abstract}
    
\vspace{-3mm}
\section{Introduction}
\label{sec:intro}

\begin{figure}[t]
  \centering
  \captionsetup{width=0.88\columnwidth}
  \setlength{\tabcolsep}{2pt}
  \renewcommand{\arraystretch}{0.5}
  \begin{tabular}{cccc}
    & \small\textbf{Original} & \small\textbf{ROI Mask} & \small\textbf{Iris Crop} \\[4pt]

    \rotatebox{90}{\small\textbf{~~~~~~~Clear}} &
    \includegraphics[height=2.05cm]{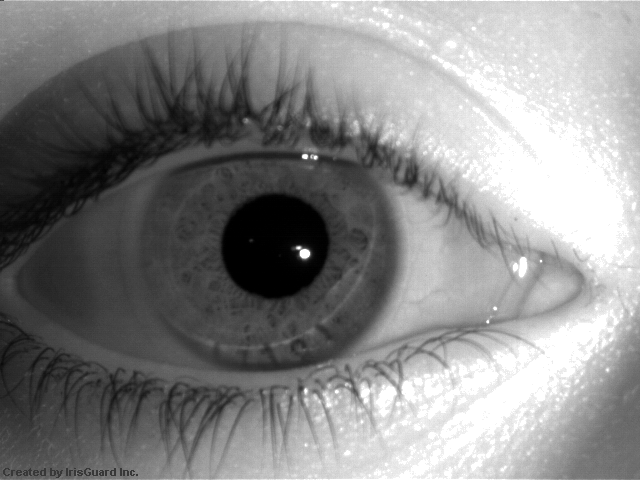} &
    \includegraphics[height=2.05cm]{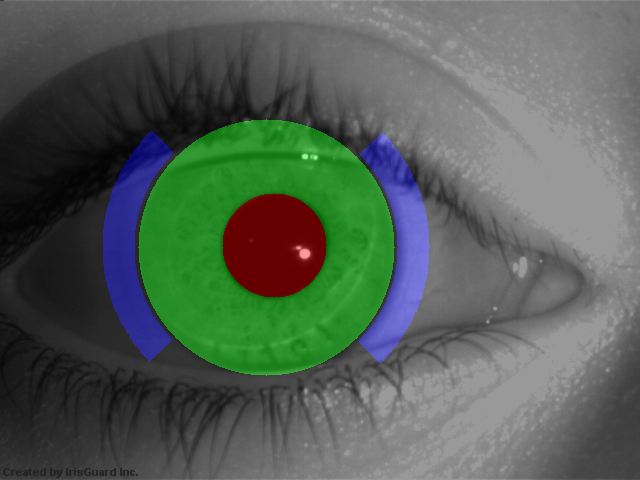} &
    \includegraphics[height=2.05cm]{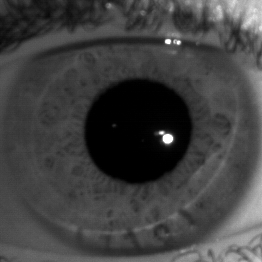} \\[1pt]

    \rotatebox{90}{\small\textbf{~~~~~~Patterned}} &
    \includegraphics[height=2.05cm]{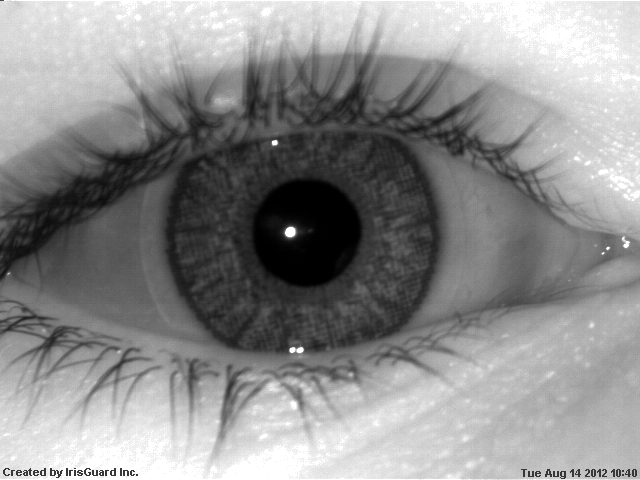} &
    \includegraphics[height=2.05cm]{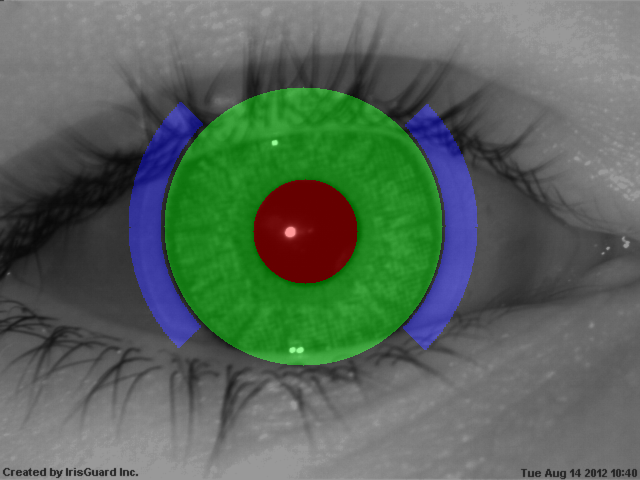} &
    \includegraphics[height=2.05cm]{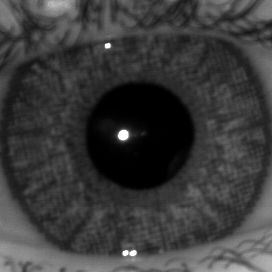} \\[1pt]

    \rotatebox{90}{\small\textbf{~~~~~~Normal}} &
    \includegraphics[height=2.05cm]{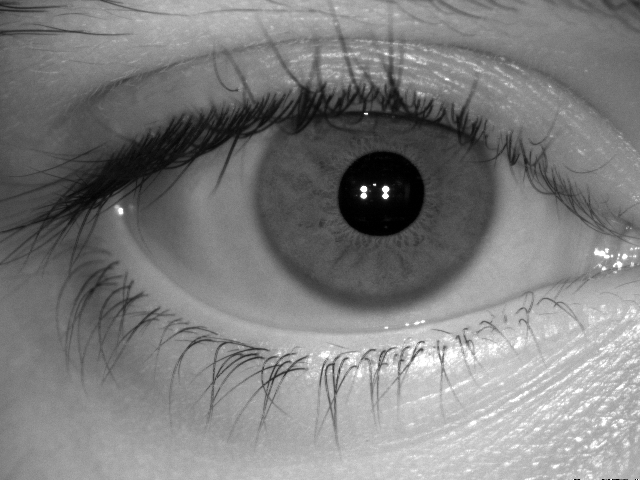} &
    \includegraphics[height=2.05cm]{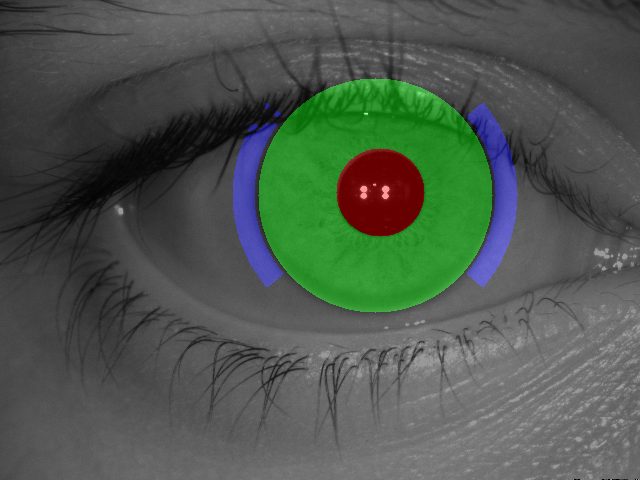} &
    \includegraphics[height=2.05cm]{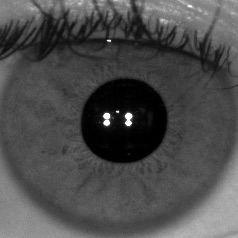} \\
  \end{tabular} 
  \caption{Sample inputs to the proposed pipeline for each lens class. Each row shows the full periocular image, the anatomical ROI mask $M$ (Eq.~\ref{eq:mask}), and the localised iris crop $\hat{I}$ used by Stage~1. In the mask column: \textbf{red} = pupil region, \textbf{green} = iris annulus, \textbf{blue} = outer arc beyond the limbus where clear-lens evidence appears.}
  \label{fig:samples}
\end{figure}
The iris is the annular region of the eye surrounding the pupil and is a powerful biometric cue whose intricate texture can even be used to distinguish identical twins~\cite{daugman2001epigenetic, daugman2004iris}.
However, this efficacy depends on an implicit assumption: that the captured iris is an unobstructed, unaltered view of bona fide ocular tissue. Contact lenses--including patterned cosmetic lenses and the far more prevalent clear\footnote{In the iris presentation attack detection (PAD) literature, clear contact lenses are also commonly referred to as transparent or soft contact lenses~\cite{erdogan2013automatic}.} lenses--challenge this assumption in clinically distinct ways (samples can be seen in Fig.\ref{fig:samples}). This is a widespread concern, as more than 140 million people worldwide were estimated to wear contact lenses in 2025~\cite{snsinsider2026contactlensmarket}.
Patterned lenses are a well-studied threat.
Cosmetic (patterned) contact lenses {\em typically} overlay a bold\footnote{Although most cosmetic contact lenses studied in the iris biometrics literature have a ``strong" textured pattern, there are other such lenses with subtle texture patterns.} synthetic texture over the iris, corrupting the Gabor-phase feature codes used for iris comparison and constituting a well-documented presentation attack~\cite{boyd2023comprehensive, hoffman2018cnn_iris_pad, yambay2023review, boyd2020iris, yadav2021cit, yadav2025multi}.
They are detectable precisely because the foreign texture is often visually conspicuous: a compact, well-aligned iris crop reveals the periodic pattern, and even lightweight classifiers can learn to separate it from a natural iris with high reliability~\cite{menotti2015deep, hoffman2018cnn_iris_pad, gupta2021generalized, yadav2014unraveling}. 
On the other hand, clear lenses are an underexplored challenge. Clear contact lenses--worn by people for vision correction--have attracted far less attention in the iris literature, on the assumption that an optically clear lens leaves the iris appearance unchanged. This assumption does not always hold in practice. Baker et al.~\cite{baker2010degradation} first showed, across more than 12{,}000 images, that clear lenses produce false non-matches through subtle deformation and edge artifacts at the iris limbus. Later work also confirmed that this degradation is not random noise but a systematic, condition-dependent bias: the refractive index mismatch at the lens boundary creates a predictable reflectance deviation in the periocular ring~\cite{doyle2013variation, yadav2019unraveling}. Later, another paper~\cite{raghavendra2017contlensnet}, proposed ContlensNet,  a method to detect clear contact lenses in ocular images.

Despite over a decade of research, detecting clear contact lenses still remains an open problem~\cite{nguyen2024deep}.
The difficulty arises from a combination of factors that together make it a challenging fine-grained classification task in biometrics:
\begin{itemize}[leftmargin=*, itemsep=1pt, topsep=1pt, parsep=0pt, partopsep=0pt]
  \item Optical invisibility:
        Clear lenses carry no colour or texture signature, unlike cosmetic contact lenses.
        The only physical evidence is a faint reflectance deviation at the iris--sclera limbus, invisible to the naked eye and easily confused with natural periocular variation.

  \item Extreme spatial localization:
        Clear lens evidence is often confined to a narrow annular band at the lens edge. Standard classification backbones trained to aggregate global information tend to ignore this small region in favour of more salient but uninformative background features.

  \item High intra-class variability:
        Different lens brands, geometries, and moisture levels produce different limbal signatures, further modulated by eye colour, pupil dilation, and illumination angle.

      \item Two fundamentally different sub-problems under one label: Patterned and clear lenses share very less commonality. A detection model trained on the bold periodic textures of a cosmetic lens is suboptimal for clear lenses.

  \item Sensor and acquisition variability:
        Iris sensors differ in resolution, near-infrared wavelength, and depth of field, all of which alter the appearance of the lens edge and surrounding periocular tissue across devices.

\end{itemize}

Our approach, motivated by the fundamental disparity between the two detection
sub-problems, i.e., cosmetic patterned lens detection and clear lens detection, utilizes a two-stage method pipeline.
Stage~1 uses an existing iris PAD model known as D-NetPAD~\cite{dnetpad} to screen for patterned lens artefacts from a compact iris crop and passing only non-patterned images to Stage~2.
Stage~2 introduces a novel Mask-Guided Attention Network that injects an anatomical ROI mask derived from classical Hough-based iris localization~\cite{daugman2004iris}, with no learned segmentation and no pixel-level annotation, directly into the backbone's multi-level feature representation.
The mask encodes the hard spatial prior that all clear lens evidence is often confined to the iris annulus and the periocular arc at the limbal boundary. Fusing this prior multiplicatively into the feature representation forces the network to commit to the anatomically correct region, eliminating the shortcut of latching onto uninformative background cues. A learned spatial gate within the same module further modulates which parts of the masked region are most diagnostic, while a Squeeze-and-Excitation block~\cite{hu2018squeeze} recalibrates feature channels globally--together asking \emph{where} to look and \emph{what} to look for.
We implement Stage~2 with three backbone architectures to assess how discriminative capacity and receptive field size interact with mask guidance: EfficientNet-B4~\cite{tan2019efficientnet},
ResNet-101~\cite{he2016deep}, and ConvNeXt-Base~\cite{liu2022convnet}. Once a clear contact lens is detected, we apply a z-score calibration for adjusting verification scores.

This work makes the following contributions:

\noindent 1. Quantitative evidence that clear lenses degrade iris verification. We quantify, across four large-scale datasets using a commercial iris-matching SDK (VeriEye)~\cite{neurotechnology_verieye}, the systematic genuine-score drop introduced by clear lenses.

\noindent 2. A two-stage method for contact-lens detection. Decomposes the three-class classification problem (``normal", ``clear", ``patterned") into a texture sub-problem (Stage~1, patterned vs. non-patterned) and a spatial localization
sub-problem (Stage~2, clear vs.\ normal), substantially outperforming single-stage baselines.

\noindent 3. A novel mask-guided attention mechanism. Injects an annotation-free anatomical ROI mask into the intermediate feature maps of a deep backbone. This combines a learned spatial attention gate with a fixed anatomy-based prior, allowing the contact-lens detector to focus on the most relevant iris regions during feature learning. Moreover, a backbone study comparing EfficientNet-B4, ResNet-101, and ConvNeXt-Base under identical mask-guidance conditions, showing that large-kernel depthwise convolutions best resolve the subtle limbal evidence of clear lens wear.

\noindent 4. Downstream verification improvement. Upto 28.3\%
relative EER reduction via predicted-label score calibration.

\section{Related Work}
\label{sec:related}


\noindent\textbf{Iris Presentation Attack Detection:}
Iris PAD has been an active area of research since the discovery that printed irides, replayed videos, and textured contact lenses can successfully fool commercial matchers~\cite{czajka2019iris, ross2019some}.
Early methods extracted hand-crafted descriptors, viz., BSIF, LBP, and
co-occurrence matrices, from the iris annulus and fed them to shallow classifiers~\cite{agarwal2023misclassifications}.
Deep learning substantially raised the bar: Menotti
et al.~\cite{menotti2015deep} showed that convolutional networks
fine-tuned end-to-end on iris crops outperform handcrafted pipelines across multiple attack types.
D-NetPAD~\cite{dnetpad} built on DenseNet-121~\cite{huang2017densely} with interpretability
constraints, demonstrated strong generalization to cosmetic lens artifacts across sensors~\cite{livdet2020}; we adopt it as our Stage~1 patterned-lens filter.
More recent work has explored multi-task and attention-based formulations.
Chen and Ross~\cite{chen2021attention_guided_iris_pad} proposed an attention-guided PAD framework that localizes discriminative iris regions using channel and spatial attention gates. The LivDet-Iris competition series~\cite{yambay2017livdet, livdet2020,tinsley2023iris} provides standardised cross-sensor benchmarks; the 2023 edition introduced GAN-generated irides as an attack category, highlighting the growing diversity of the threat landscape.
The recent top-performing LivDet-Iris competition~\cite{mitcheff2025iris} achieved good performance, showing that PAD is a well-studied problem for patterned lenses. 

\noindent\textbf{Clear Lens Detection:}
Unlike patterned contact lenses, clear lenses leave no strong texture
signature, making detection fundamentally harder.
Baker et al.~\cite{baker2010degradation} established the empirical
baseline, demonstrating that clear lenses increase false non-match rates even for high-quality NIR sensors.
Doyle and Bowyer~\cite{doyle2013variation} later showed that the lens
boundary introduces a measurable 3D surface distortion detectable with photometric stereo, though this approach requires controlled multi-light acquisition. Raghavendra et al.~\cite{raghavendra2017contlensnet} proposed ContlensNet, a 15-layer CNN that classifies normalized iris patches as no lens, soft lens, or textured lens and determines the image-level label through majority voting. Yadav et al.~\cite{yadav2019unraveling} conducted a large-scale analysis of the effect of textured and incidentally clear lenses on commercial matchers, quantifying the score drop across multiple cohorts. Unlike ContlensNet~\cite{raghavendra2017contlensnet}, which only reports lens classification performance, our work goes beyond that by reporting iris recognition and score calibration results. Further,  we use the pre-normalized ocular image, which includes the scleral region, thereby capturing the outer boundary of lenses. 

\noindent\textbf{Attention in Deep Recognition Models:}
Attention mechanisms have become a standard tool for directing network capacity to informative spatial regions~\cite{hu2018squeeze, woo2018cbam}.
The Squeeze-and-Excitation (SE) block~\cite{hu2018squeeze} introduced channel-wise recalibration via global average pooling, allowing the network to selectively amplify informative feature channels.
CBAM~\cite{woo2018cbam} extended this with a sequential spatial gate, computing a 2D attention map from both average- and max-pooled features.
Non-local networks~\cite{wang2018non} and Vision
Transformers~\cite{dosovitskiy2021vit} model long-range dependencies via self-attention, but at a cost in data efficiency on small biometric datasets.
In the medical imaging domain, mask-guided attention networks have been proposed to incorporate anatomical segmentations as spatial priors, improving sensitivity to lesion regions without requiring lesion-level supervision~\cite{jafrasteh2024mga}.
Our approach is inspired by this line of work: rather than learning attention maps entirely from the training signal, we inject a Hough-derived iris ROI mask as a hard prior directly into the backbone's intermediate feature hierarchy, coupling it with a learned spatial gate and a SE recalibration block.
This is, to our knowledge, the first application of anatomy-driven external mask guidance to contact-lens detection.

\noindent\textbf{Backbone Architectures:} ResNet~\cite{he2016deep} established deep residual learning as the standard baseline for visual recognition, and it remains competitive on biometric tasks due to its simplicity and well-understood inductive bias.
EfficientNet~\cite{tan2019efficientnet} introduced compound coefficient scaling of width, depth, and resolution, achieving strong accuracy-efficiency trade-offs on ImageNet while keeping parameter counts low attractive for resource-constrained biometric deployment.
ConvNeXt~\cite{liu2022convnet} revisited the ResNet design space in light of Vision Transformer innovations, replacing $3\!\times\!3$ convolutions with $7\!\times\!7$ depthwise kernels, adopting LayerNorm, and applying the inverted major challenge structure; the resulting architecture matches or exceeds Swin Transformer on standard benchmarks while remaining fully convolutional.
For our task, the enlarged receptive field of ConvNeXt is especially
relevant: the limbal evidence of a clear lens spans a narrow but spatially extended annular band, and larger kernels are better positioned to capture the subtle curvature and reflectance gradients that distinguish a lens boundary from natural periocular tissue.
We compare all three backbones under identical mask-guidance conditions to isolate the contribution of receptive field size to clear-lens discriminability.


\section{Proposed Methodology}
\label{sec:method}
\subsection{Pipeline Overview}

\begin{figure}[t]
  \centering
  \includegraphics[width=\columnwidth]{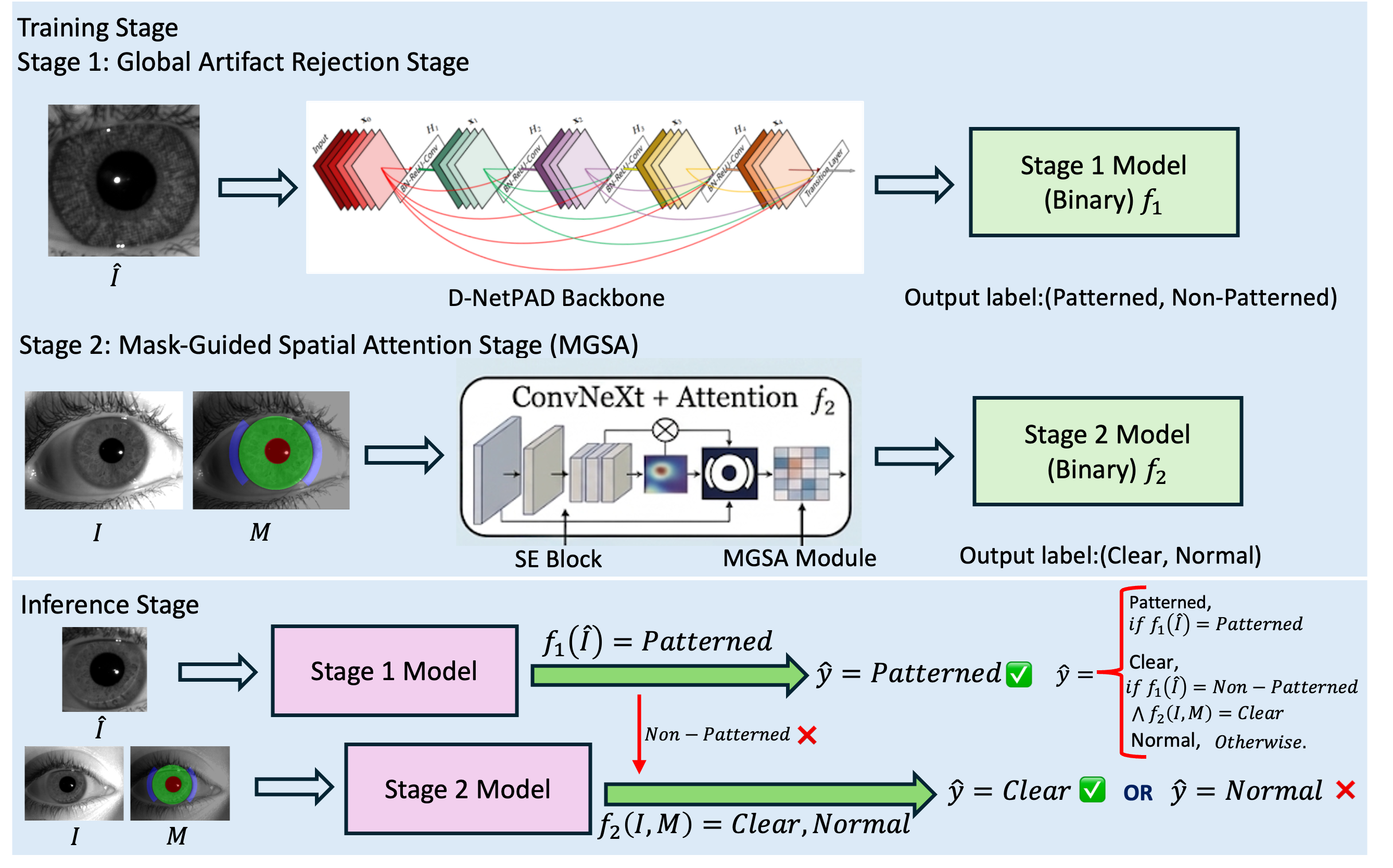} 
 \caption{Two-Stage Contact Lens Detection Framework.
    Top: Stage~1 uses cropped iris images to detect patterned lenses, while Stage~2 uses full periocular images and ROI masks to classify clear vs.\ normal lenses.
    Bottom: At inference, images predicted as non-patterned by Stage~1 are passed to Stage~2 for final classification. $I$ denotes the full iris image, $M$ denotes the corresponding mask, and $\hat{I}$ denotes the cropped iris image.}
  \label{fig:two_stage_pipeline}
\end{figure}

Contact lens detection is not a single problem; it is two fundamentally different problems sharing the same output space. Patterned lenses introduce a strong global texture anomaly detectable from a compact iris crop, while clear lenses leave only a faint annular trace in the vicinity of the limbal boundary.
Treating both with a single three-class classifier forces the network to simultaneously learn coarse texture suppression and subtle annular contrast differences, which compete for representational capacity and degrade performance on the harder class. We address this asymmetry with a two-stage method that routes each image through two specialized binary decisions:
\vspace{-2mm}
\begin{equation}
\hat{y} =
\begin{cases}
\texttt{patterned},
& \text{if } f_1(\hat{I}) = \texttt{patterned}, \\[3pt]

\texttt{clear},
& \begin{aligned}[t]
   \text{if } &\ f_1(\hat{I}) = \texttt{non-patterned} \\
              &\land\ f_2(I,M) = \texttt{clear},
  \end{aligned} \\[6pt]

\texttt{normal},
& \text{otherwise.}
\end{cases}
\label{eq:method}
\end{equation}
where, $\hat{I}$ is a localized iris crop fed to Stage~1, $I$ is the full periocular image fed to Stage~2, and $M$ is an automatically derived anatomical ROI mask. $f_1 : \hat{I} \to {\texttt{patterned},\ \texttt{non-patterned}}$ is the Stage~1 D-NetPAD binary classifier, and $f_2 : (I, M) \to {\texttt{clear},\ \texttt{normal}}$ is the Stage~2 ConvNeXt-Base + MGSA classifier.
Positive Stage~1 detections exit immediately; only non-patterned images reach the computationally heavier Stage~2, keeping the clear-lens sub-problem free from easy cases that would otherwise dominate the gradient. Fig.~\ref{fig:two_stage_pipeline} illustrates the complete pipeline.

\subsection{Iris Masks}
\label{sec:masks}

The fundamental spatial prior for clear-lens detection is that all evidence is confined to the iris and the narrow band immediately surrounding it at the limbus.
We encode this prior as a binary mask $M \in \{0,1\}^{H \times W}$ derived entirely from classical iris geometry so that no learned segmentation and no pixel-level annotation are required.
For pupil and limbus localization, we apply the circular Hough transform~\cite{shah2009iris} twice to each greyscale image. The circular Hough transform~\cite{shah2009iris, quinn2024open} is a voting-based method that finds circles in an edge image by accumulating evidence for all $(x, y, r)$ triples consistent with detected edge points. For the pupil, the image is first median-blurred and thresholded to isolate the dark pupil disk; Canny edges are extracted and fed to \texttt{HoughCircles}; the accumulator threshold is lowered iteratively until at least ten candidate circles are returned. The final pupil circle $(c_x^p,\,c_y^p,\,r_p)$--where $c_x^p, c_y^p$ are the horizontal and vertical coordinates of the pupil centre and $r_p$ is its radius--is the coordinate-wise mean of the accepted candidates, which suppresses outliers from eyelash occlusion.
The limbus circle $(c_x^\ell,\,c_y^\ell,\,r_\ell)$ is found with the same multi-scale search, constrained so that its centre lies within $0.25\,r_p$ of the pupil centre and its radius exceeds
$1.5\,r_p$. Three-region labelling.
Given the two detected circles, every pixel $\mathbf{p} = (x,y)$ is assigned one of three anatomical labels:
\begin{equation}
\begingroup
\setlength{\jot}{0pt}
\ell(\mathbf{p}) =
\begin{cases}
\texttt{pupil}, & \text{if } \|\mathbf{p}-\mathbf{c}^p\|_2 \le r_p, \\[-1pt]
\texttt{iris}, & \text{if } r_p < \|\mathbf{p}-\mathbf{c}^{\ell}\|_2 \le r_{\ell}, \\[-1pt]
\texttt{outer\_arc}, &
  \begin{aligned}[t]
  \text{if } &\ r_{\ell} < \|\mathbf{p}-\mathbf{c}^{\ell}\|_2 \le r_{\ell}+\delta \\
             &\land\ \theta(\mathbf{p}) \in \Theta,
  \end{aligned} \\[-1pt]
0, & \text{otherwise.}
\end{cases}
\endgroup
\label{eq:labelling}
\end{equation}





The \texttt{outer\_arc} region captures the periocular tissue immediately beyond the limbus, where the lens edge creates its most visible reflectance deviation.
It is restricted to the inferior and superior quadrants
$\Theta = [45^\circ, 135^\circ] \cup [225^\circ, 315^\circ]$
(measured from the horizontal), since eyelid occlusion dominates the
nasal and temporal sectors and provides no lens cue.
The radial offset $\delta = 35$\,px beyond the detected limbus accommodates localization uncertainty.
The final binary mask is the union of all three labelled regions:
\vspace{-2mm}
\begin{equation}
  M(\mathbf{p}) =
  \mathbf{1}\!\left[\ell(\mathbf{p}) \in
    \{\texttt{pupil},\;\texttt{iris},\;\texttt{outer\_arc}\}\right].
  \label{eq:mask}
\end{equation}
Fig.~\ref{fig:samples} shows representative masks for each lens class. However, the generated masks are not always robust, leaving room for further improvement. Since mask quality directly affects the final classification decision, accurate mask generation is crucial to the overall performance of the proposed method.
\subsection{Stage 1 — Patterned Lens Screening}
\label{sec:stage1}

Stage~1 solves a binary classification problem: does the image contain a patterned (cosmetic) lens?
Patterned lenses impose a bold, periodic synthetic texture over the iris that is immediately apparent in a well-aligned iris crop $\hat{I}$ of size $256 \times 256$.
We adopt D-NetPAD~\cite{dnetpad}, a DenseNet 121~\cite{huang2017densely} fine-tuned specifically for iris presentation attack detection, as our Stage~1 classifier, $f_1$.
D-NetPAD is pre-trained on a large pool of iris PAD data and generalizes well to the cosmetic lens texture anomaly; we fine-tune only its classification head on each target dataset.
Images predicted as patterned exit the pipeline immediately with label \texttt{patterned}.
Images not predicted as patterned are passed to Stage~2, which now operates on a cleaner distribution free of the dominant texture cue.
\subsection{Stage 2 — Mask-Guided Attention Network}
\label{sec:stage2}
Stage 2 addresses the truly challenging sub-problem: distinguishing a bare iris (\texttt{normal}) from one wearing a clear lens (\texttt{clear}).
Unlike patterned lenses, which alter the global iris texture, clear lenses preserve it for the most part. The only physical evidence of their presence is a faint reflectance discontinuity confined to the narrow annular band at the iris limbus.
Two design decisions follow directly from this observation.
First, we use the \emph{full} periocular image $I$ at its native resolution ($480 \times 640$) rather than a tight iris crop, to preserve the spatial extent of the limbal region.
Second, we introduce a Mask-Guided Spatial Attention (MGSA) module that injects the anatomical mask $M$ directly into the backbone's intermediate feature hierarchy, preventing the model from attending to periocular background structures that carry no lens signal.
\subsubsection{Backbone}
\label{sec:backbone}

We instantiate Stage~2 with ConvNeXt-Base~\cite{liu2022convnet} pre-trained on ImageNet-1K~\cite{deng2009imagenet}.
The ConvNeXt architecture consists of a stem followed by four stages with progressively downsampled feature maps; channel dimensions for the Base variant are $[128, 256, 512, 1024]$ at stages 0--3 respectively.
We inject MGSA after Stage~2 of the backbone ($C{=}512$ channels), where the feature map retains sufficient spatial resolution to resolve the limbal cues while encoding semantically meaningful mid-level representations.
Following the attention module, a global average pool and a dropout layer ($p{=}0.4$) precede the two-class linear head.
We compare ConvNeXt-Base against ResNet-101~\cite{he2016deep} and
EfficientNet-B4~\cite{tan2019efficientnet} under identical MGSA conditions in our ablation study (Sec.~\ref{sec:results}).

\subsubsection{Mask-Guided Spatial Attention (MGSA)}
\label{sec:mgsa}

Let $\bm{F} \in \mathbb{R}^{B \times C \times H' \times W'}$ be the feature map produced after the chosen backbone stage.
The core insight driving MGSA is that two complementary signals are needed to localize the lens cue: (i)~\emph{where} to look, encoded by the anatomical mask $M$, and (ii)~\emph{what to look for}, encoded by a learned spatial saliency gate.
Neither signal alone is sufficient: the mask without learning provides no discriminative weighting within the iris, and a learned gate without the mask may attend to eyelid or periocular skin features that correlate with the label only due to dataset biases. The mask is first binarized and upsampled by nearest-neighbour interpolation to match the feature-map resolution:
\begin{equation}
  \tilde{M} = \mathrm{Upsample}(M,\,(H',W'))
  \;\in\;\{0,1\}^{H' \times W'}.
  \label{eq:upsample}
\end{equation}
A learned saliency gate is produced by a single $1{\times}1$ convolution that collapses the channel dimension to a scalar confidence map:
\begin{equation}
  G = \sigma\!\bigl(\mathbf{w}_g \ast \bm{F}\bigr)
  \;\in\;(0,1)^{1 \times H' \times W'},
  \quad \mathbf{w}_g \in \mathbb{R}^{1 \times C \times 1 \times 1}.
  \label{eq:gate}
\end{equation}
The MGSA output is the element-wise triple product:
\begin{equation}
  \boxed{
  \bm{F}^* = \bm{F} \;\odot\; G \;\odot\; \tilde{M},
  }
  \label{eq:mgsa}
\end{equation}
where $\odot$ denotes spatially broadcast element-wise multiplication.
$\tilde{M}$ acts as a \emph{hard} anatomical gate: it unconditionally zeros all activations outside the iris region, regardless of what the network has learned.
$G$ provides a \emph{soft} learned re-weighting of positions within that constrained region, learning to further up-weight the outer annulus and limbal arc where the lens boundary appears.

SE Channel Recalibration.
After spatial gating, a Squeeze-and-Excitation (SE)
block~\cite{hu2018squeeze} recalibrates $\bm{F}^*$ channel-wise with
reduction ratio $r{=}16$:
\begin{equation}
  \bm{F}^{**} = \bm{F}^* \;\odot\;
    \sigma\!\Bigl(\mathbf{W}_2\,
      \mathrm{ReLU}\!\bigl(\mathbf{W}_1\,
        \mathrm{GAP}(\bm{F}^*)\bigr)\Bigr),
  \label{eq:se}
\end{equation}
where $\mathrm{GAP}$ denotes global average pooling. MGSA gates the spatial (H×W) dimensions of the C-channel feature map, zeroing activations outside the iris annulus uniformly across all channels; SE then reweights those same C channels globally, selecting which feature types encode the lens signal most strongly. The two are complementary: MGSA determines where to look; SE determines what to look for within that region.

\subsection{Training and Testing}
\label{sec:training}
\textbf{Subject-stratified cross-validation.}
Biometric datasets exhibit strong intra-subject correlation: all images of the same identity share iris texture, colour, and periocular appearance independent of lens type.
Naive random splitting allows the same subject to appear on both sides of a fold boundary, enabling the model to exploit identity-specific cues rather than lens appearance---a form of \emph{identity leakage} that inflates validation accuracy and masks poor generalization. We eliminate this with subject-stratified 5-fold cross-validation.
(i)~\emph{identity disjointness}---all images of the same subject are confined to a single fold; and (ii)~\emph{class-balance preservation}---the three-class distribution is maintained across folds.
The held-out test split (20\% of subjects, never seen during training or model selection) is fixed at the dataset level. The best model checkpoint for each fold is selected by balanced accuracy on that fold's held-out training subjects. Here ``checkpoint'' denotes the saved model weights from the best-performing epoch during training. ``Held-out training subjects" are the subjects kept aside from the training fold as the validation data to decide which epoch is best --- this is different from the final test set. ``Balanced accuracy" is used instead of just ``accuracy'' because the two classes (clear vs. normal) may not have equal numbers of images. Balanced accuracy averages the per-class recall, so a model that ignores the minority class cannot score well.

\textbf{Ensemble inference.} At test time, the five independently trained Stage-2 models $\{f_k\}_{k=1}^5$ are combined by averaging their softmax posterior distributions:
\vspace{-1mm}
\begin{equation}
  \hat{p}(c\mid I,M)
  = \frac{1}{5}\sum_{k=1}^{5} p_k(c\mid I,M),
  \;
  \hat{y} = \operatorname*{arg\,max}_{c}\hat{p}(c\mid I,M).
  \label{eq:ensemble}
\end{equation}
Since each fold is trained on a disjoint subject population, the five models are complementary in the identities they have observed; averaging their posteriors reduces decision-boundary variance without any additional computational cost at training time. For the optimization, stage~2 is trained with AdamW~\cite{loshchilov2019decoupled} ($\lambda{=}10^{-4}$), cosine annealing ($\eta_0{=}2{\times}10^{-4}$, $\eta_{\min}{=}10^{-6}$, $T{=}80$ epochs), cross-entropy loss with label smoothing ($\varepsilon{=}0.05$), and dropout ($p{=}0.4$) before the classification head.
Stage~1 is fine-tuned with SGD ($\eta_0{=}5{\times}10^{-4}$, step decay $\times0.1$ every 10 epochs, 50 epochs total, weight decay $10^{-4}$).

\subsection{Lens-informed Score Calibration}
\label{sec:calibration}

Once the two-stage classifier assigns a label to every enrolled and probe image, we apply a per-condition z-score calibration to the raw VeriEye verification scores. For each pair type $t \in \{\texttt{NN},\,\texttt{CC},\,\texttt{CN}\}$, we estimate the mean $\mu_t$ and standard deviation $\sigma_t$ from the \emph{genuine} pairs of that type in the training set.
Every score is then rescaled onto the Normal-Normal (NN) reference distribution:
\vspace{-2mm}
\begin{equation}
  s_{\text{cal}} =
    \frac{s - \mu_t}{\sigma_t}\,\sigma_{\texttt{NN}} + \mu_{\texttt{NN}},
  \label{eq:calib}
\end{equation}
where, $(\mu_{\texttt{NN}},\,\sigma_{\texttt{NN}})$ are the parameters of the NN reference distribution of the training set.
This mapping is a structure-preserving affine transform: it shifts the mean of each non-NN distribution onto the NN mean and matches its spread, closing the score gap introduced by the lens without altering the rank ordering within any single pair-type group.
Crucially, the pair type for each verification pair is determined entirely by the classifier's \emph{predicted} labels---no ground-truth lens information is used at inference time.\footnote{We also performed calibration using ground-truth lens labels and obtained the same accuracy, confirming that the predicted labels were sufficient for this evaluation.}

\FloatBarrier
\section{Experiments and Results}
\label{sec:results}
\subsection{Contact Lens Classification}
\label{subsec:cls}

\definecolor{headerblue}{RGB}{220,230,242}
\definecolor{subheaderblue}{RGB}{235,242,250}
\definecolor{avggray}{RGB}{245,245,245}

\definecolor{headerblue}{RGB}{220,230,242}
\definecolor{subheaderblue}{RGB}{235,242,250}
\definecolor{avggray}{RGB}{245,245,245}

\begin{table*}[htbp]
\centering
\caption{Two-stage contact-lens detection results across all four datasets.
Stage~1 (D-NetPAD) classifies patterned vs.\ non-patterned irises.
Stage~2 (ConvNeXt-Base + MGSA) classifies normal vs.\ clear irises on the non-patterned subset; therefore, Stage~2 accuracy is computed only on non-patterned test images.
Accuracy $\pm$ values denote bootstrap standard errors from 10,000 resamples.
Train/test splits are subject-stratified with no identity overlap.
\textsuperscript{*}Approximately 50 images with apparent labeling inconsistencies were excluded from the UND and IITD datasets during preprocessing; however, these inconsistencies were not exhaustively verified across the full datasets.
\textsuperscript{b}Baseline CCR (\%) reproduced from ContlensNet~\cite{raghavendra2017contlensnet} using intra-sensor validation from Table~1. ContlensNet was trained and tested using its own IIITD/ND protocol and a different number of images; therefore, its results are not directly comparable to the subject-stratified splits used here.}
\label{tab:results}

\renewcommand{\arraystretch}{1.0}
\setlength{\tabcolsep}{2.5pt}
\footnotesize

\resizebox{\textwidth}{!}{%
\begin{tabular}{
l
rrrrr
S[table-format=3.2]
ccc
rrrr
}
\toprule
\rowcolor{headerblue}
\textbf{Dataset}
& \multicolumn{5}{c}{\textbf{Dataset Statistics}\textsuperscript{*}}
& \multicolumn{4}{c}{\textbf{Accuracy (\%)}}
& \multicolumn{4}{c}{\textbf{Subject-Stratified Split}} \\

\rowcolor{subheaderblue}
& \textbf{Imgs}
& \textbf{IDs}
& \textbf{Pat.}
& \textbf{Nor.}
& \textbf{Clr.}
& {\textbf{S1}}
& \textbf{S2}
& \textbf{Pipe.}
& \textbf{ContlensNet\textsuperscript{b}}
& \textbf{Tr. IDs}
& \textbf{Tr. $N$}
& \textbf{Te. IDs}
& \textbf{Te. $N$} \\
\midrule

UND-AD100~\cite{doyle2014ndcld}
& 300 & 41 & 100 & 100 & 100
& 100.00
& $87.50 \pm 5.28$
& $\mathbf{90.00 \pm 4.23}$
& 95.00
& 32 & 250 & 9 & 50 \\

UND-LG4000~\cite{doyle2014ndcld}
& 1200 & 76 & 400 & 400 & 400
& 100.00
& $92.00 \pm 2.22$
& $\mathbf{95.20 \pm 1.35}$
& 96.91
& 61 & 950 & 15 & 250 \\

IITD-Cogent~\cite{kohli2013revisiting,yadav2014unraveling}
& 3508 & 101 & 1202 & 1163 & 1143
& 100.00
& $96.15 \pm 0.87$
& $\mathbf{97.50 \pm 0.56}$
& 86.73
& 79 & 2747 & 22 & 761 \\

IITD-Vista~\cite{kohli2013revisiting,yadav2014unraveling}
& 2906 & 101 & 1005 & 940 & 961
& 100.00
& $98.18 \pm 0.69$
& $\mathbf{98.80 \pm 0.45}$
& 87.33
& 81 & 2322 & 20 & 584 \\

\midrule
\rowcolor{avggray}
\textbf{Average}
& -- & -- & -- & -- & --
& {\bfseries 100.00}
& $\mathbf{93.46}$
& $\mathbf{95.38}$
& $91.49$
& -- & -- & -- & -- \\
\bottomrule
\end{tabular}%
}

\vspace{3pt}

\parbox{\textwidth}{%
\raggedright
\footnotesize
\emph{S1}: Stage~1;
\emph{S2}: Stage~2 evaluated on the non-patterned subset only;
\emph{Pat.}: patterned;
\emph{Nor.}: normal;
\emph{Clr.}: clear;
\emph{Pipe.}: complete pipeline;
\emph{Tr.}: training split;
\emph{Te.}: test split.
}

\end{table*}

\definecolor{trainblue}{RGB}{221,235,247}
\begin{table}[t]
\centering
\caption{Per-class correct predictions on a test split (correct\,/\,total). ``Total'' is the number of test images, and ``correct'' is the correctly classified test images.}
\label{tab:per_class_counts}
\renewcommand{\arraystretch}{1.2}
\setlength{\tabcolsep}{5pt}
\small
\begin{tabular}{l c c c c}
\toprule
\rowcolor{headerblue}
\textbf{Dataset} & \textbf{Patterned} & \textbf{Normal}
                 & \textbf{Clear}     & \textbf{Overall} \\
\midrule
UND-AD100
  & \makecell{10/10\\(100.0\%)} & \makecell{16/20\\(80.0\%)}
  & \makecell{19/20\\(95.0\%)}  & \makecell{\textbf{45/50}\\\textbf{(90.00\%)}} \\[2pt]
\midrule  
UND-LG4000
  & \makecell{100/100\\(100.0\%)} & \makecell{80/80\\(100.0\%)}
  & \makecell{58/70\\(82.9\%)}   & \makecell{\textbf{238/250}\\\textbf{(95.20\%)}} \\[2pt]
\midrule  
IITD-Cogent
  & \makecell{268/268\\(100.0\%)} & \makecell{240/243\\(98.8\%)}
  & \makecell{234/250\\(93.6\%)} & \makecell{\textbf{742/761}\\\textbf{(97.50\%)}} \\[2pt]
\midrule  
IITD-Vista
  & \makecell{200/200\\(100.0\%)} & \makecell{192/192\\(100.0\%)}
  & \makecell{185/192\\(96.4\%)} & \makecell{\textbf{577/584}\\\textbf{(98.80\%)}} \\
\bottomrule
\end{tabular}
\end{table}

\begin{table}[t]
\caption{VeriEye EER (\%) before and after score calibration.
Calibration parameters are estimated from genuine and impostor pairs; 
the presence or absence of a lens is determined by the proposed two-stage 
classifier (real-world protocol, no oracle labels used).}
\label{tab:recognition}
\centering
\small
\renewcommand{\arraystretch}{1.2}

\begin{tabular*}{\columnwidth}{@{\extracolsep{\fill}}lccc@{}}
\toprule
\textbf{Dataset}
  & \textbf{Before (\%)}
  & \textbf{After (\%)}
  & \textbf{Rel.\ Red.} \\
\midrule
UND-AD100   & 0.49 & 0.47 & $-4.1\%$  \\
UND-LG4000  & 0.75 & 0.64 & $-14.7\%$ \\
IITD-Cogent & 0.76 & 0.70 & $-7.9\%$  \\
IITD-Vista  & 2.26 & 1.62 & $-28.3\%$ \\
\bottomrule
\end{tabular*}
\end{table}
Table~\ref{tab:results} reports Stage-1, Stage-2, and full pipeline accuracy together with per-class scores across all four datasets. Table~\ref{tab:per_class_counts} provides the per-class correct-prediction breakdown.
Stage-1 achieves perfect patterned detection.
Across all four datasets, Stage-1 (D-NetPAD) correctly classifies every patterned iris in the test set — 100\% accuracy with zero bootstrap SD (Table~\ref{tab:per_class_counts}).
Patterned lenses introduce salient colour and periodic texture artefacts that are easily separable from natural iris textures, confirming that binary patterned/non-patterned classification is a well-conditioned problem for a standard convolutional network fine-tuned from ImageNet weights. Clear lens detection is the harder sub-problem.
The main classification challenge lies in Stage-2.
Across all datasets, the Clear lens accuracy is the lowest of the three classes, confirming that clear lens detection is harder than patterned detection and also harder than normal iris detection. As shown in Table~\ref{tab:results}, the gap is most pronounced on AD100 (S2$_\text{Acc.}$\,=\,87.50\%) and LG4000 (S2$_\text{Acc.}$\,=\,92.00), which are the smallest datasets: with fewer training subjects, the model has less opportunity to learn the subtle reflectance deviation at the limbal boundary that distinguishes a clear lens from a bare iris. MGSA drives good Stage-2 accuracy on large datasets.
On the two large IITD datasets, the MGSA-equipped ConvNeXt-Base achieves 96.15\% (Cogent) and 98.18\% (Vista) Stage-2 accuracy.
By hard-zeroing all activations outside the iris and periocular arc via the anatomical mask $\tilde{M}$, the network cannot exploit background structures; the soft learned gate $G$ then refines attention within that constrained region to the specific annular sub-areas where lens evidence physically appears. For cross-sensor analysis, despite substantial differences in sensor type, image resolution, and subject demographics across the four datasets, the pipeline achieves 90.0\%--98.8\% accuracy with a bootstrap SD between 0.45 and 4.23. This cross-sensor consistency supports the generalization of the proposed two-stage method.

\subsection{Score and Calibration}
\label{subsec:calibration}

\begin{figure*}[t]
  \centering
  \includegraphics[width=0.245\linewidth]{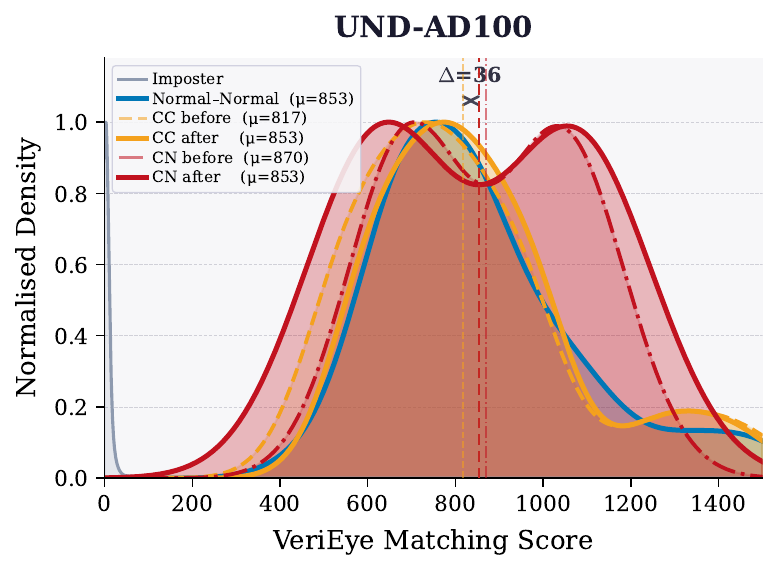}
  \includegraphics[width=0.245\linewidth]{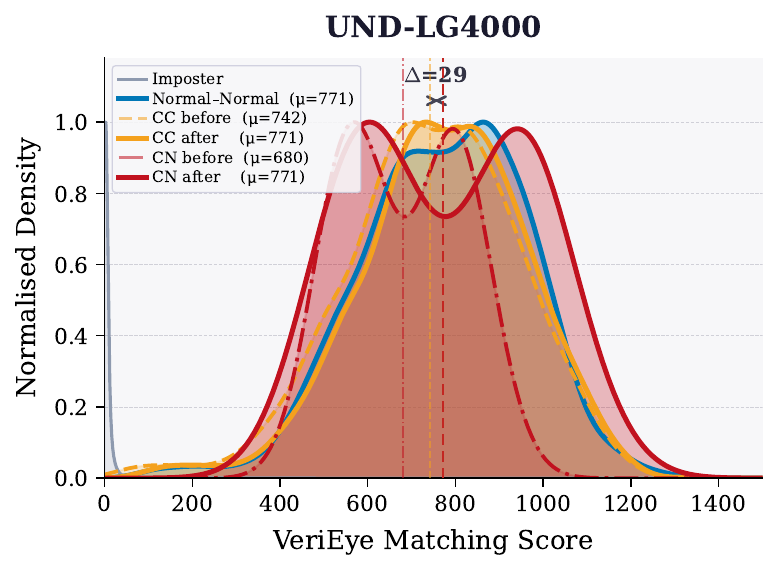}
  \includegraphics[width=0.245\linewidth]{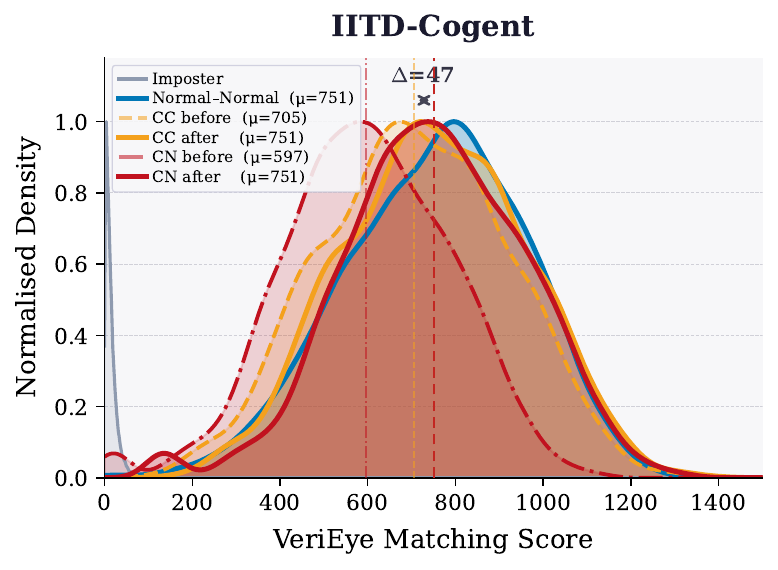}
  \includegraphics[width=0.245\linewidth]{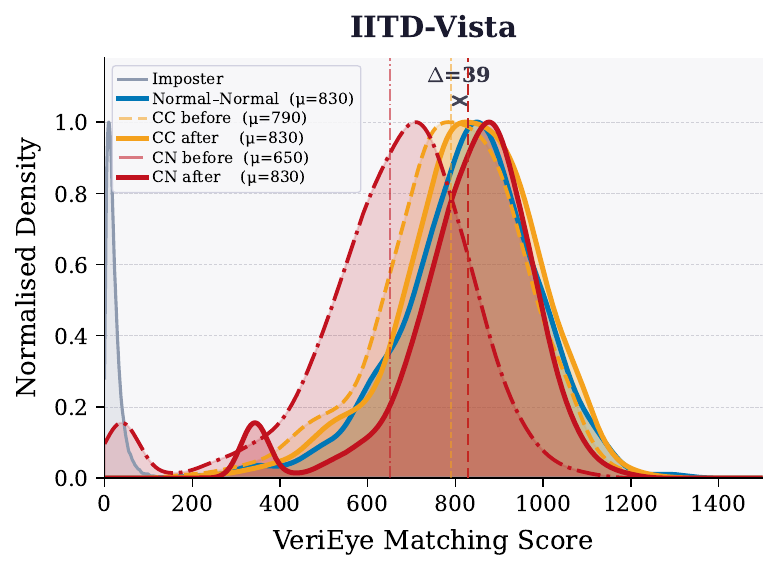}
  \caption{VeriEye genuine-pair score distributions before (dashed/faded) and after (solid) lens-informed score calibration, for all four datasets.
  \textbf{Blue}: Normal--Normal (reference, unchanged).
  \textbf{Amber}: Clear--Clear before $\to$ after.
  \textbf{Red}: Clear--Normal before $\to$ after.
  The $\Delta$ annotation marks the mean gap between Normal--Normal and Clear--Clear before calibration.
  Pair types are determined by the two-stage classifier predictions (no ground-truth labels used).}
  \label{fig:score_degradation}
\end{figure*}

\begin{figure}[t]
  \centering
  \includegraphics[width=1.0\linewidth]{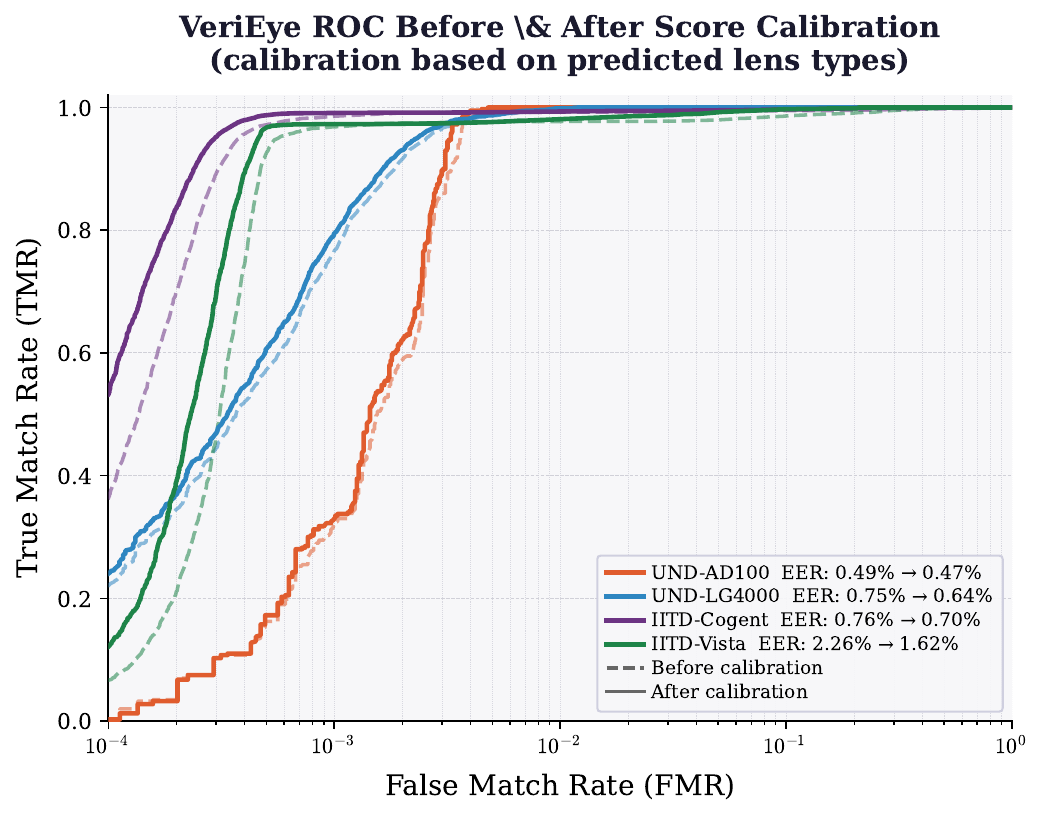}
  \caption{VeriEye ROC curves before (dashed) and after (solid) score calibration for all four datasets.
  EER values are annotated in the legend.
  Calibration uses pair types predicted by the proposed two-stage classifier under a real-world protocol
  (no oracle labels, 100\% pair coverage).}
  \label{fig:roc}
\end{figure}

Clear lenses cause systematic score degradation.
Fig.~\ref{fig:score_degradation} visualizes the VeriEye genuine-pair score distributions for each pair type before calibration.
In every dataset, Clear-Clear (CC) genuine pairs score measurably lower than Normal-Normal (NN) pairs on the same subjects.
The score gap ranges from 36 points on AD100 to 39 points on IITD-Vista, consistent with the systematic bias introduced by the refractive index mismatch at the lens boundary.
Clear--Normal (CN) cross-condition pairs exhibit an even larger displacement on IITD-Vista ($\mu_\text{CN}$\,=\,650 vs.\ $\mu_\text{NN}$\,=\,830), pushing genuine match scores closer to the impostor distribution (grey curve, concentrated near zero).
This confirms that clear lenses introduce a \emph{predictable, condition-dependent} bias in iris matching--not random noise--and that the bias is present across all four sensors and cohorts tested. Calibration closes the gap.
After applying the z-score calibration of Eq.~(\ref{eq:calib}) with pair types predicted by the two-stage classifier, the CC and CN distributions shift to align with the NN reference (Fig.~\ref{fig:score_degradation}, solid curves).
On IITD-Vista, the CN mean rises from 650 to 830, completely closing the 180-point gap.
On AD100, the CC distribution (amber solid) overlaps almost perfectly with the NN reference (blue), confirming that the calibration parameters estimated from genuine pairs generalise correctly to all pairs in the verification database.
Crucially, this improvement is achieved using only the \emph{predicted} lens types from the two-stage classifier--no ground-truth label information is available at inference time. Verification EER improves across all datasets.
Table~\ref{tab:recognition} and Fig.~\ref{fig:roc} report EER before and after calibration.
The improvement is consistent across all four datasets, ranging from a modest 4.1\% relative reduction on AD100 (0.49\%\,$\to$\,0.47\%) to a substantial
28.30\% on IITD-Vista (2.26\%\,$\to$\,1.62\%).
The magnitude of the EER improvement correlates with the severity of the initial score degradation: IITD-Vista has the largest gap (39 points) and the largest CN displacement, and accordingly shows the greatest benefit from
calibration.
AD100, with a smaller $\Delta$ (36 points) and a small number of non-patterned pairs, shows the most modest improvement. The ROC curves in Fig.~\ref{fig:roc} confirm that the improvement is not confined to the EER operating point: the calibrated curve (solid) lies above the raw curve (dashed) across the full range of False Match Rates, from $10^{-4}$ to $10^{0}$.
This means that regardless of the security threshold chosen by the operator, calibrated scores yield a higher TMR than raw scores.

\subsection{Backbone Comparison}
\label{subsec:backbone}

\begin{table}[t]
\centering
\caption{Stage-2 backbone comparison: full pipeline accuracy (\%) with standard deviations from 10,000 resamples. Stage~1 is fixed.}
\label{tab:backbone}
\renewcommand{\arraystretch}{1.12}
\setlength{\tabcolsep}{4pt}
\small
\resizebox{\columnwidth}{!}{%
\begin{tabular}{lccc}
\toprule
\rowcolor{headerblue}
\textbf{Dataset}
& \textbf{EfficientNet-V2 + MGSA}
& \textbf{ResNet-101 + MGSA}
& \textbf{ConvNeXt-B + MGSA} \\
\midrule
UND-AD100   & $58.00 \pm 6.99$ & $90.00 \pm 4.26$          & $\mathbf{90.00 \pm 4.26}$ \\
UND-LG4000  & $72.00 \pm 2.83$ & $\mathbf{95.60 \pm 1.29}$ & $95.20 \pm 1.38$ \\
IITD-Cogent & $68.20 \pm 1.67$ & $97.24 \pm 0.60$          & $\mathbf{97.50 \pm 0.56}$ \\
IITD-Vista  & $65.41 \pm 1.96$ & $98.29 \pm 0.54$          & $\mathbf{98.80 \pm 0.45}$ \\
\bottomrule
\end{tabular}%
}
\end{table}
Table~\ref{tab:backbone} compares three Stage-2 backbone architectures. ResNet-101, EfficientNet-V2-S, and ConvNeXt-Base all incorporate MGSA spatial attention. MGSA substantially outperforms the mask-free baseline. We conducted experiments on EfficientNet-V2-S without mask guidance, which degraded by up to 30\% on AD100 (60.00\% vs.\ 90.00\%) and by 10 points on IITD-Cogent (87.25\% vs.\ 97.50\%), confirming that the anatomical mask is the primary driver of Stage-2 performance, not backbone capacity alone. ConvNeXt-Base is the best MGSA backbone.
Under identical mask guidance, ConvNeXt-Base matches or exceeds ResNet-101 on three of four datasets; ResNet-101 holds a marginal edge on LG4000 (+0.40 percentage points (pp)). On the larger IITD datasets, ConvNeXt-Base leads by +0.26 pp on Cogent and +0.51 pp on Vista, indicating that its $7{\times}7$ depthwise kernels better capture the subtle limbal texture evidence of clear lens wear.
The $7{\times}7$ depthwise kernels of ConvNeXt capture the spatially extended limbal arc more effectively than ResNet's $3{\times}3$ convolutions, and this advantage is most pronounced where there is sufficient data to exploit the larger receptive field.
\subsection{Cross-Dataset Generalization}
\label{subsec:cross}                                                                                        
\begin{table}[t]
\centering
\caption{Cross-dataset pipeline accuracy (\%). Rows = train dataset; columns = test dataset.
$^\dagger$Cross-sensor only (shared subjects).}
\label{tab:cross}
\renewcommand{\arraystretch}{1.15}
\setlength{\tabcolsep}{2.2pt}
\footnotesize
\begin{tabular}{lcccc}
\toprule
\rowcolor{headerblue}
\textbf{Train\,{\scriptsize$\downarrow$} Test\,{\scriptsize$\rightarrow$}} & \textbf{UND-AD100} & \textbf{UND-LG4000} & \textbf{IITD-Cogent} & \textbf{IITD-Vista} \\
\midrule
UND-AD100    & {---}  & 87.60          & 69.12           & 36.99 \\
UND-LG4000   & 68.00  & {---}          & 61.63           & 43.66 \\
IITD-Cogent  & 78.00  & \textbf{97.20} & {---}           & $^\dagger$\textbf{97.26} \\
IITD-Vista   & 84.00  & 73.20          & $^\dagger$85.41 & {---} \\
\bottomrule
\end{tabular}
\end{table}
Table~\ref{tab:cross} reports pipeline accuracy for all 12 source$\to$target combinations using the same checkpoints as Table~\ref{tab:results}, with no fine-tuning on the target dataset.   
The results show that the proposed method generalizes well within the same sensor family and moderately across related sensors, with the IITD-Cogent model achieving 97.20\% on UND-LG4000 and 97.26\% on IITD-Vista, suggesting that the anatomical mask provides a robust, sensor-invariant spatial prior. Even under cross-family transfer, IITD-trained models remain well above chance, indicating that MGSA captures meaningful limbal cues rather than purely sensor-specific patterns. In contrast, models trained on the smaller UND datasets generalize poorly to IITD sensors, highlighting an asymmetric transfer gap likely caused by limited training diversity and substantial differences in image resolution and depth of field, and motivating future work on domain adaptation and stronger data augmentation.

  

\section{Conclusion}
\label{sec:conclusion}

We presented a two-stage method for iris contact lens detection that exploits the visual asymmetry between patterned and clear lenses. Stage~1 addresses the easier problem of patterned lens detection from cropped iris images with high accuracy across all four datasets, while Stage~2 addresses the harder clear vs. normal classification problem using a ConvNeXt-Base model with the proposed MGSA, which combines an anatomical ROI mask, spatial gating, and channel recalibration to focus on the most informative iris and periocular regions. Across UND-AD100, UND-LG4000, IITD-Cogent, and IITD-Vista, the full pipeline achieves 90.0\%--98.8\% accuracy and bootstrap SD scores, with ConvNeXt-Base consistently outperforming EfficientNet and ResNet baselines. We further showed that clear lenses systematically degrade commercial iris matcher scores, pushing genuine scores closer to the impostor distribution, and that condition-aware z-score calibration based on the predicted lens labels can recover much of this loss, reducing EER by 4.1\%--28.3\% across all datasets. Although the current system relies on Hough-based anatomical masks and has been evaluated only on near-infrared data, the results demonstrate that lens-aware detection and calibration provide a practical path toward more robust real-world iris recognition.
\vspace{+2mm}

\noindent\textbf{Acknowledgments:} This work was supported by NSF CITeR under Awards \#1841517 and \#2413309.  The code supporting the findings of this study is publicly available on GitHub \href{https://github.com/iPRoBe-lab/MGSA_ContactLens_Detection}
{[https://github.com/iPRoBe-lab/MGSA\_ContactLens\_Detection]}.

{\small
\bibliographystyle{IEEEtran}
\bibliography{main}

@article{baker2010degradation,
  title={Degradation of iris recognition performance due to non-cosmetic prescription contact lenses},
  author={Baker, Stephen and Bowyer, Kevin W and Flynn, Patrick J},
  journal={Computer Vision and Image Understanding},
  volume={114},
  number={9},
  pages={1030--1044},
  year={2010},
  publisher={Elsevier}
}

@article{menotti2015deep,
  title   = {Deep Representations for Iris, Face, and Fingerprint Spoofing Detection},
  author  = {Menotti, David and Chiachia, Gibran and Pinto, Alberto and Schwartz, William Robson and Pedrini, Helio and Falc{\~a}o, Alexandre X. and Rocha, Anderson},
  journal = {IEEE Transactions on Information Forensics and Security},
  volume  = {10},
  number  = {4},
  pages   = {864--879},
  year    = {2015},
  doi     = {10.1109/TIFS.2015.2398817}
}

@inproceedings{huang2017densely,
  author    = {Huang, Gao and Liu, Zhuang and van der Maaten, Laurens and Weinberger, Kilian Q.},
  title     = {Densely Connected Convolutional Networks},
  booktitle = {Proceedings of the IEEE/CVF Conference on Computer Vision and Pattern Recognition (CVPR)},
  pages     = {4700--4708},
  year      = {2017},
  doi       = {10.1109/CVPR.2017.243}
}

@article{daugman2004iris,
  author    = {Daugman, John},
  title     = {How Iris Recognition Works},
  journal   = {IEEE Transactions on Circuits and Systems for Video Technology},
  volume    = {14},
  number    = {1},
  pages     = {21--30},
  year      = {2004},
  doi       = {10.1109/TCSVT.2003.818350}
}

@article{yadav2019unraveling,
  author    = {Yadav, Daman and Kohli, Naman and Doyle, James S. and Singh, Richa and Vatsa, Mayank and Bowyer, Kevin W.},
  title     = {Unraveling the Effect of Textured Contact Lenses on Iris Recognition},
  journal   = {IEEE Transactions on Information Forensics and Security},
  volume    = {9},
  number    = {5},
  pages     = {851--862},
  year      = {2014},
  doi       = {10.1109/TIFS.2014.2313025}
}

@inproceedings{dnetpad,
  title={{D-NetPAD}: An explainable and interpretable iris presentation attack detector},
  author={Sharma, Renu and Ross, Arun},
  booktitle={IEEE International Joint Conference on Biometrics (IJCB)},
  pages={1--10},
  year={2020},
  organization={}
}

@inproceedings{he2016deep,
  author    = {He, Kaiming and Zhang, Xiangyu and Ren, Shaoqing and Sun, Jian},
  title     = {Deep Residual Learning for Image Recognition},
  booktitle = {Proceedings of the IEEE Conference on Computer Vision and Pattern Recognition (CVPR)},
  pages     = {770--778},
  year      = {2016},
  doi       = {10.1109/CVPR.2016.90}
}

@inproceedings{liu2022convnet,
  author    = {Liu, Zhuang and Mao, Hanzi and Wu, Chao-Yuan and Feichtenhofer, Christoph and Darrell, Trevor and Xie, Saining},
  title     = {A {ConvNet} for the 2020s},
  booktitle = {Proceedings of the IEEE/CVF Conference on Computer Vision and Pattern Recognition (CVPR)},
  pages     = {11976--11986},
  year      = {2022},
  doi       = {10.1109/CVPR52688.2022.01167}
}

@inproceedings{tan2019efficientnet,
  author    = {Tan, Mingxing and Le, Quoc V.},
  title     = {{EfficientNet}: Rethinking Model Scaling for Convolutional Neural Networks},
  booktitle = {Proceedings of the International Conference on Machine Learning (ICML)},
  pages     = {6105--6114},
  year      = {2019}
}

@inproceedings{hu2018squeeze,
  author    = {Hu, Jie and Shen, Li and Sun, Gang},
  title     = {Squeeze-and-Excitation Networks},
  booktitle = {Proceedings of the IEEE/CVF Conference on Computer Vision and Pattern Recognition (CVPR)},
  pages     = {7132--7141},
  year      = {2018},
  doi       = {10.1109/CVPR.2018.00745}
}

@inproceedings{woo2018cbam,
  author    = {Woo, Sanghyun and Park, Jongchan and Lee, Joon-Young and Kweon, In So},
  title     = {{CBAM}: Convolutional Block Attention Module},
  booktitle = {Proceedings of the European Conference on Computer Vision (ECCV)},
  pages     = {3--19},
  year      = {2018},
  doi       = {10.1007/978-3-030-01234-2_1}
}

@article{dosovitskiy2021vit,
  author    = {Dosovitskiy, Alexey and Beyer, Lucas and Kolesnikov, Alexander and Weissenborn, Dirk and Zhai, Xiaohua and Unterthiner, Thomas and Dehghani, Mostafa and Minderer, Matthias and Heigold, Georg and Gelly, Sylvain and Uszkoreit, Jakob and Houlsby, Neil},
  title     = {An Image is Worth 16x16 Words: Transformers for Image Recognition at Scale},
  journal   = {International Conference on Learning Representations (ICLR)},
  year      = {2021}
}

@inproceedings{yambay2017livdet,
  author    = {Yambay, David and Becker, Blaine and Kohli, Naman and Yadav, Daman and Czajka, Adam and Bowyer, Kevin W. and Schuckers, Stephanie and Singh, Richa and Vatsa, Mayank and Noore, Afzel and Gragnaniello, Diego and Sansone, Carlo and Verdoliva, Luisa},
  title     = {{LivDet} Iris 2017 — Iris Liveness Detection Competition 2017},
  booktitle = {Proceedings of the IEEE International Joint Conference on Biometrics (IJCB)},
  pages     = {733--741},
  year      = {2017},
  doi       = {10.1109/BTAS.2017.8272763}
}

@inproceedings{livdet2020,
  author    = {Das, Priyanka and McGrath, Joseph and Fang, Zhaoyuan and Parzianello, Luciano and Czajka, Adam},
  title     = {Iris Liveness Detection Competition ({LivDet-Iris}) — The 2020 Edition},
  booktitle = {Proceedings of the IEEE International Joint Conference on Biometrics (IJCB)},
  pages     = {1--10},
  year      = {2020},
  doi       = {10.1109/IJCB48548.2020.9304890}
}

@article{nguyen2024deep,
  title={Deep learning for iris recognition: A survey},
  author={Nguyen, Kien and Proen{\c{c}}a, Hugo and Alonso-Fernandez, Fernando},
  journal={ACM Computing Surveys},
  volume={56},
  number={9},
  pages={1--35},
  year={2024},
  publisher={ACM New York, NY}
}

@inproceedings{hoffman2018cnn_iris_pad,
  title={Convolutional Neural Networks for Iris Presentation Attack Detection: Toward Cross-Dataset and Cross-Sensor Generalization},
  author={Hoffman, Steven and Sharma, Renu and Ross, Arun},
  booktitle={Proceedings of the IEEE Conference on Computer Vision and Pattern Recognition Workshops},
  year={2018}
}

@inproceedings{chen2021attention_guided_iris_pad,
  title={An Explainable Attention-Guided Iris Presentation Attack Detector},
  author={Chen, Cunjian and Ross, Arun},
  booktitle={Proceedings of the IEEE/CVF Winter Conference on Applications of Computer Vision Workshops},
  pages={97--106},
  year={2021}
}

@inproceedings{gupta2021generalized,
  title={Generalized iris presentation attack detection algorithm under cross-database settings},
  author={Gupta, Mehak and Singh, Vishal and Agarwal, Akshay and Vatsa, Mayank and Singh, Richa},
  booktitle={25th International Conference on Pattern Recognition (ICPR)},
  pages={5318--5325},
  year={2021}
}

@inproceedings{doyle2013variation,
  title={Variation in accuracy of textured contact lens detection based on sensor and lens pattern},
  author={Doyle, James S and Bowyer, Kevin W and Flynn, Patrick J},
  booktitle={IEEE Sixth International Conference on Biometrics: Theory, applications and Systems (BTAS)},
  pages={1--7},
  year={2013},
  organization={}
}

@inproceedings{agarwal2023misclassifications,
  title={Misclassifications of contact lens iris {PAD} algorithms: Is it gender bias or environmental conditions?},
  author={Agarwal, Akshay and Ratha, Nalini and Noore, Afzel and Singh, Richa and Vatsa, Mayank},
  booktitle={Proceedings of the IEEE/CVF Winter Conference on Applications of Computer Vision},
  pages={961--970},
  year={2023}
}

@article{jafrasteh2024mga,
  title={MGA-Net: A novel mask-guided attention neural network for precision neonatal brain imaging},
  author={Jafrasteh, Bahram and Lubi{\'a}n-L{\'o}pez, Sim{\'o}n Pedro and Trimarco, Emiliano and Ruiz, Macarena Roman and Barrios, Carmen Rodr{\'\i}guez and Almagro, Yolanda Mar{\'\i}n and Benavente-Fern{\'a}ndez, Isabel},
  journal={NeuroImage},
  volume={300},
  pages={120872},
  year={2024},
  publisher={Elsevier}
}

@inproceedings{czajka2019iris,
  title={Iris presentation attack detection based on photometric stereo features},
  author={Czajka, Adam and Fang, Zhaoyuan and Bowyer, Kevin},
  booktitle={IEEE Winter Conference on Applications of Computer Vision (WACV)},
  pages={877--885},
  year={2019},
  organization={}
}

@inproceedings{wang2018non,
  title={Non-local neural networks},
  author={Wang, Xiaolong and Girshick, Ross and Gupta, Abhinav and He, Kaiming},
  booktitle={Proceedings of the IEEE Conference on Computer Vision and Pattern Recognition},
  pages={7794--7803},
  year={2018}
}

@misc{neurotechnology_verieye,
  author       = {{Neurotechnology}},
  title        = {{VeriEye SDK}},
  howpublished = {\url{https://www.neurotechnology.com/verieye.html}},
  note         = {Iris identification technology and software development kit},
  year         = {2026},
  accessed     = {2026-04-27}
}

@inproceedings{deng2009imagenet,
  title     = {ImageNet: A Large-Scale Hierarchical Image Database},
  author    = {Deng, Jia and Dong, Wei and Socher, Richard and Li, Li-Jia and Li, Kai and Fei-Fei, Li},
  booktitle = {Proceedings of the IEEE Conference on Computer Vision and Pattern Recognition},
  pages     = {248--255},
  year      = {2009}
}

@inproceedings{loshchilov2019decoupled,
  title     = {Decoupled Weight Decay Regularization},
  author    = {Loshchilov, Ilya and Hutter, Frank},
  booktitle = {International Conference on Learning Representations},
  year      = {2019}
}

@techreport{doyle2014ndcld,
  title       = {Notre Dame Image Database for Contact Lens Detection in Iris Recognition--2013},
  author      = {Doyle, Jay and Bowyer, Kevin W.},
  institution = {University of Notre Dame, Computer Vision Research Laboratory},
  year        = {2014},
  note        = {Dataset documentation/README}
}

@inproceedings{kohli2013revisiting,
  title     = {Revisiting Iris Recognition with Color Cosmetic Contact Lenses},
  author    = {Kohli, Naman and Yadav, Daksha and Vatsa, Mayank and Singh, Richa},
  booktitle = {Proceedings of the International Conference on Biometrics},
  pages     = {1--7},
  year      = {2013},
  doi       = {10.1109/ICB.2013.6613021}
}

@article{yadav2014unraveling,
  title   = {Unraveling the Effect of Textured Contact Lenses on Iris Recognition},
  author  = {Yadav, Daksha and Kohli, Naman and Doyle, James S. and Singh, Richa and Vatsa, Mayank and Bowyer, Kevin W.},
  journal = {IEEE Transactions on Information Forensics and Security},
  volume  = {9},
  number  = {5},
  pages   = {851--862},
  year    = {2014},
  doi     = {10.1109/TIFS.2014.2313025}
}

@misc{snsinsider2026contactlensmarket,
  author       = {{S\&S Insider}},
  title        = {{Contact Lens Market Size, Share \& Growth Report 2035}},
  year         = {2026},
  howpublished = {\url{https://www.snsinsider.com/reports/contact-lens-market-8566}},
  note         = {Last updated: March 2, 2026. Accessed: April 30, 2026}
}

@article{boyd2023comprehensive,
  title={Comprehensive study in open-set iris presentation attack detection},
  author={Boyd, Aidan and Speth, Jeremy and Parzianello, Lucas and Bowyer, Kevin W and Czajka, Adam},
  journal={IEEE Transactions on Information Forensics and Security},
  volume={18},
  pages={3238--3250},
  year={2023},
  publisher={IEEE}
}

@incollection{yambay2023review,
  title={Review of iris presentation attack detection competitions},
  author={Yambay, David and Das, Priyanka and Boyd, Aidan and McGrath, Joseph and Fang, Zhaoyuan and Czajka, Adam and Schuckers, Stephanie and Bowyer, Kevin and Vatsa, Mayank and Singh, Richa and others},
  booktitle={Handbook of Biometric Anti-Spoofing: Presentation Attack Detection and Vulnerability Assessment},
  pages={149--169},
  year={2023},
  publisher={Springer}
}

@article{boyd2020iris,
  title={Iris presentation attack detection: Where are we now?},
  author={Boyd, Aidan and Fang, Zhaoyuan and Czajka, Adam and Bowyer, Kevin W},
  journal={Pattern Recognition Letters},
  volume={138},
  pages={483--489},
  year={2020},
  publisher={Elsevier}
}

@inproceedings{erdogan2013automatic,
  title={Automatic detection of non-cosmetic soft contact lenses in ocular images},
  author={Erdogan, Gizem and Ross, Arun},
  booktitle={Biometric and Surveillance Technology for Human and Activity Identification X},
  volume={8712},
  pages={62--76},
  year={2013},
  organization={SPIE}
}

@article{shah2009iris,
  title={Iris segmentation using geodesic active contours},
  author={Shah, Samir and Ross, Arun},
  journal={IEEE Transactions on Information Forensics and Security},
  volume={4},
  number={4},
  pages={824--836},
  year={2009},
  publisher={IEEE}
}

@article{quinn2024open,
  title={An Open Source Iris Segmentation Algorithm for Non-ideal Images},
  author={Quinn, George W},
   journal={NIST},
  year={2024},
  publisher={George W. Quinn}
}

@inproceedings{yadav2021cit,
  title={{CIT-GAN}: Cyclic image translation generative adversarial network with application in iris presentation attack detection},
  author={Yadav, Shivangi and Ross, Arun},
  booktitle={Proceedings of the IEEE/CVF Winter Conference on Applications of Computer Vision},
  pages={2412--2421},
  year={2021}
}

@inproceedings{yadav2025multi,
  title={A Multi-domain Image Translative Diffusion StyleGAN for Iris Presentation Attack Detection},
  author={Yadav, Shivangi and Ross, Arun},
  booktitle={Proceedings of the IEEE/CVF International Conference on Computer Vision Workshop},
  pages={3688--3697},
  year={2025}
}

@article{daugman2001epigenetic,
  title={Epigenetic randomness, complexity and singularity of human iris patterns},
  author={Daugman, John and Downing, Cathryn},
  journal={Proceedings of the Royal Society of London. Series B: Biological Sciences},
  volume={268},
  number={1477},
  pages={1737--1740},
  year={2001},
  publisher={The Royal Society}
}

@inproceedings{ross2019some,
  title={Some research problems in biometrics: The future beckons},
  author={Ross, Arun and Banerjee, Sudipta and Chen, Cunjian and Chowdhury, Anurag and Mirjalili, Vahid and Sharma, Renu and Swearingen, Thomas and Yadav, Shivangi},
  booktitle={International Conference on Biometrics (ICB)},
  pages={1--8},
  year={2019},
  organization={}
}

@inproceedings{tinsley2023iris,
  title={Iris liveness detection competition (livdet-iris)--the 2023 edition},
  author={Tinsley, Patrick and Purnapatra, Sandip and Mitcheff, Mahsa and Boyd, Aidan and Crum, Colton and Bowyer, Kevin and Flynn, Patrick and Schuckers, Stephanie and Czajka, Adam and Fang, Meiling and others},
  booktitle={IEEE International Joint Conference on Biometrics (IJCB)},
  pages={1--10},
  year={2023},
  organization={}
}

@inproceedings{raghavendra2017contlensnet,
  title={Contlensnet: Robust iris contact lens detection using deep convolutional neural networks},
  author={Raghavendra, Ramachandra and Raja, Kiran B and Busch, Christoph},
  booktitle={2017 IEEE winter conference on applications of computer vision (WACV)},
  pages={1160--1167}
}

@inproceedings{mitcheff2025iris,
  title={Iris Liveness Detection Competition (LivDet-Iris)--The 2025 Edition},
  author={Mitcheff, Mahsa and Hossain, Afzal and Webster, Samuel and Khan, Siamul and Roszczewska, Katarzyna and Tapia, Juan E and Stockhardt, Fabian and Gonz{\'a}lez-Soler, L{\'a}zaro J and Lim, Ji-Young and Pollok, Mirko and others},
  booktitle={IEEE International Joint Conference on Biometrics (IJCB)},
  pages={1--10},
  year={2025}
}
}

\end{document}